\documentclass[10pt,twocolumn,letterpaper]{article}

\usepackage{iccv}
\usepackage{times}
\usepackage{epsfig}
\usepackage{graphicx}
\usepackage{amsmath}
\usepackage{amssymb}
\usepackage[T1]{fontenc}
\usepackage{mathrsfs}
\usepackage{booktabs}
\usepackage{multirow}
\usepackage{array}
\usepackage{colortbl}

\usepackage[pagebackref=true,breaklinks=true,letterpaper=true,colorlinks,bookmarks=false]{hyperref}

\iccvfinalcopy 

\def\iccvPaperID{****} 

\ificcvfinal\fi
\begin{document}

\title{Semi-supervised Skin Detection by Network with Mutual Guidance}


\author{ 
Yi He$^{1*}$\quad
Jiayuan Shi$^2$\thanks{Equal Contributors}\quad\ %
Chuan Wang$^2$\thanks{Corresponding Author}\quad\
Haibin Huang$^2$\quad
Jiaming Liu$^2$\\
Guanbin Li$^3$\quad\
Risheng Liu$^1$\quad\
Jue Wang$^2$ \\
{\normalsize $^1$Dalian University of Technology, } 
{\normalsize \tt \{heyi@mail.,rsliu@\}dlut.edu.cn} \\ 
{\normalsize $^2$Megvii Technology, }
{\normalsize \tt \{shijiayuan,wangchuan,huanghaibin,liujiaming,wangjue\}@megvii.com}\\ %
{\normalsize $^3$Sun Yat-sen University, }
{\normalsize \tt liguanbin@mail.sysu.edu.cn}
}

\maketitle

\begin{abstract}\label{sec:abs}
In this paper we present a new data-driven method for robust skin detection from a single human portrait image.
Unlike previous methods, we incorporate human body as a weak semantic guidance into this task, considering acquiring large-scale of human labeled skin data is commonly expensive and time-consuming. To be specific, we propose a dual-task neural network for joint detection of skin and body via a semi-supervised learning strategy. The dual-task network contains a shared encoder but two decoders for skin and body separately. For each decoder, its output also serves as a guidance for its counterpart, making both decoders mutually guided. Extensive experiments were conducted to demonstrate the effectiveness of our network with mutual guidance, and experimental results show our network outperforms the state-of-the-art in skin detection.


\end{abstract}
\section{Introduction}\label{sec:intro}
\begin{figure}[t]
  \centering
  \includegraphics[width=0.98\linewidth]{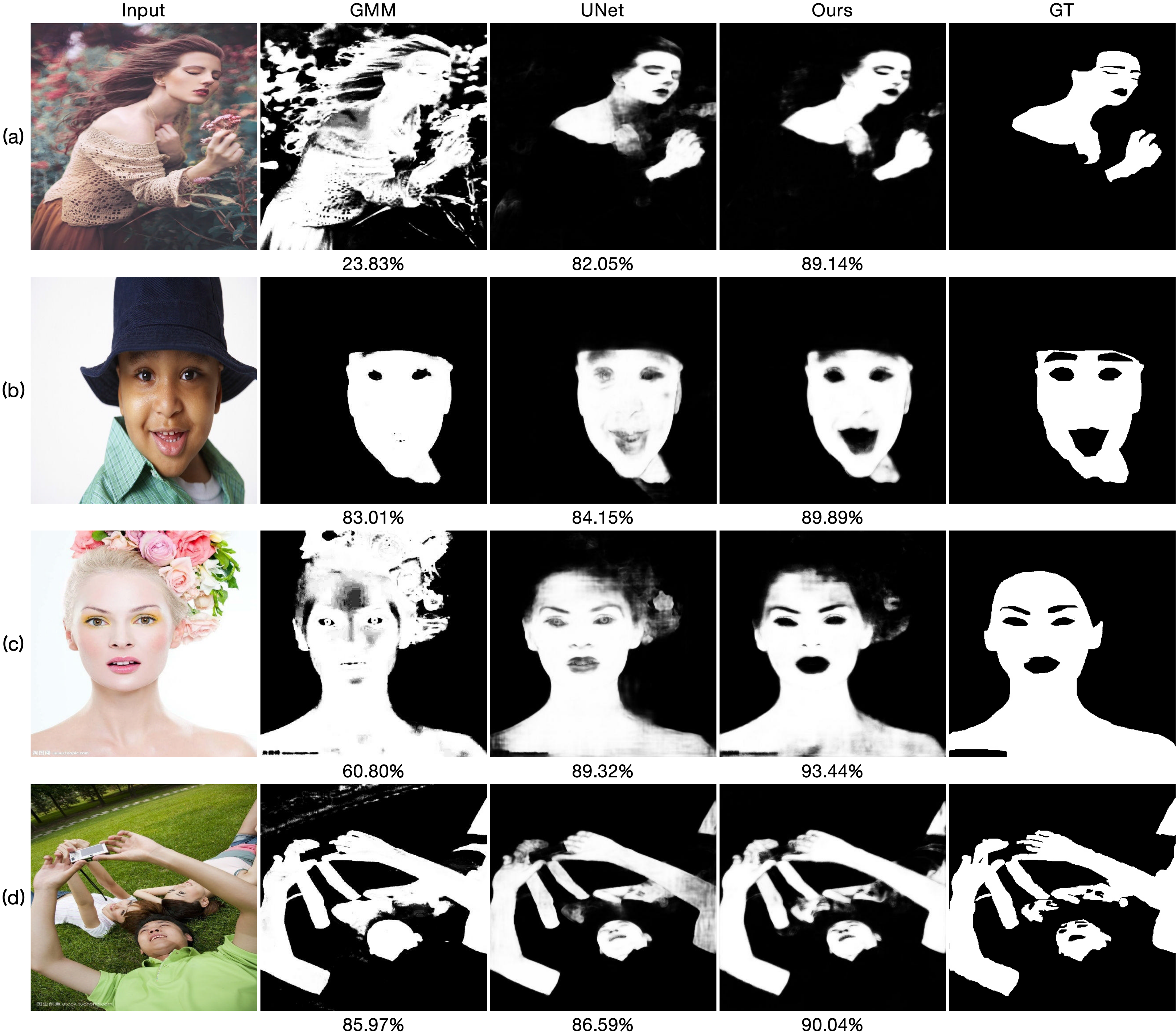}\\
  \caption{Skin detection results by our approach vs. solutions of a UNet and a tradition Gaussian Mixture Model (GMM). The intersection-over-union (IoU) rates demonstrate our approach has a better performance.}\label{fig:teaser}
\end{figure}

Skin detection is the process of finding skin-colored pixels and regions from images and videos.  It is a very interesting problem and typically serves as a pre-processing step for further applications like face detection, gestures detection, semantic filtering of web contents and so on \cite{chen2016skin,rautaray2015vision,chai2001skin,jaimes2007multimodal,fang2016computational}.

Skin detection has been proven quite challenging with large variation in skin appearance, depending on its own color proprieties and illumination conditions. Previous methods like \cite{erdem2011combining,zhu2004adaptive} tried to model skin color in different color spaces and train skin classifiers in those spaces. However, these methods heavily rely on the distribution of skin color, and with no sematic information involved, they suffer from a limited performance. In recent years, with the development of deep neural networks, skin detection methods have been proposed by adaption of networks used for other detection tasks \cite{he2016deep,chen2018encoder,ronneberger2015u}. Although these DNN based skin detection methods reveal promising accuracy improvements, they are still limited by annotated skin data which is expensive and time-consuming to collect.

\begin{figure*}[t]
    \centering
    \includegraphics[width=0.98\linewidth]{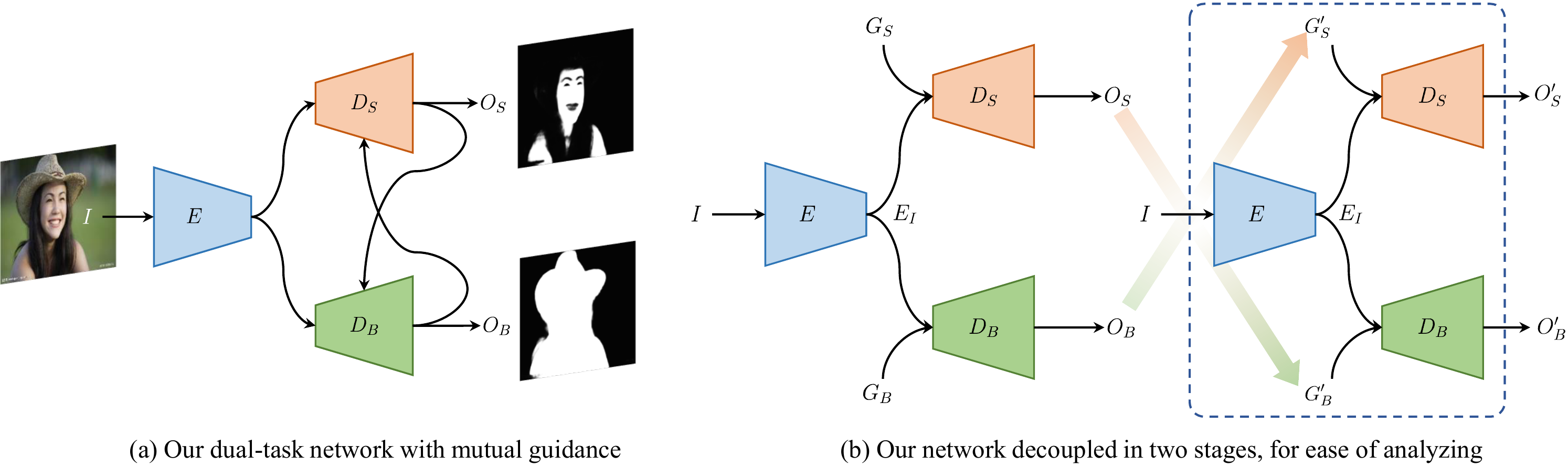}
    \caption{Structure of our dual-task network with mutual guidance. (a) The original network structure with mutual guidance loop. (b) The decoupled network into two stages for ease of analyzing.}
    \label{fig:network-structure}
\end{figure*}

To this end, we propose to improve skin detection by introducing body detection as a guidance. If a body mask is available, it could potentially facilitate the skin detection in two-folds. First, it provides a prior information for a skin detector where higher probability of skin is located. Second, after a skin mask is detected, it can filtered out the false positive pixels in the background. Meanwhile, with skin mask as a guidance, a body detector is also provided with more information. To enable the mutual guidance scheme, we designed a dual-task neural network for jointly detection of skin and body. The entire network contains a shared encoder but two decoders for skin and body detection separately. The output from each decoder would be fed to the other one so as to form a recurrent loop as shown in~\figurename~\ref{fig:network-structure}(a). The shared encoder of the two detectors would extract common feature maps from the input image, considering the similarity of the two tasks and the compactness of the network. This structure enables us to train the skin detection network without increasing the annotated training data but simply adding a human body mask dataset, which is rather easier to obtain. Since the two datasets contain two types of ground truth seperately, i.e. a data sample has either a target skin mask or body mask, we train the network in a semi-supervised manner with a newly designed loss and a customized training strategy. Experimental results demonstrate the effectiveness of all the newly involved techniques for our network, and qualitative, quantitative evaluations also show that our network outperforms state-of-the-art methods as shown in~\figurename~\ref{fig:teaser},~\ref{fig:image-results} and~\tableautorefname~\ref{tab:values} for skin detection task. We also build a new dataset composed of 5,000 annotated skin masks and 5,711 annotated body masks which can be released for future research upon the acceptance of this paper.



To summarize, our main contributions are:
\begin{itemize}
\setlength\itemsep{0.05em}
\item A novel uniform dual-task neural network with mutual guidance, for joint detection of skin and body which can boost the performance of both tasks, especially for skin.
\item A newly designed loss and customized training strategy within a semi-supervised fashion, which performs well in the case of missing ground truth information.
\item A new dataset containing skin and body annotated masks to demonstrate the effectiveness of our network, and facilitate the future research in community.
\end{itemize}
\section{Related Work}\label{sec:related_work}

\paragraph{Skin detection and segmentation.}
Skin detection has been studied in the past two decades. Existing methods can be grouped into three categories, i.e. defining boundary models explicitly on color spaces~\cite{kovac2003human,erdem2011combining,powar2013skin,qiang2010robust,tan2012gesture} (thresholding), applying traditional machine learning techniques to learn a skin color model~\cite{liu2010robust,zaidan2014multi,zhu2004adaptive}, and using a deep neural network to learn an end-to-end model for skin segmentation~\cite{al2013impact,wu2012skin,kovac2003human,seow2003neural,chen2002skin}. The thresholding methods focus on defining a specified region in color spaces like RGB, YCbCr, HSV so that a pixel falls in the regions is considered to be skin. However, there is a significant overlap between the skin and non-skin pixels in color space, for example numerous objects in the background such as wall, cloth could also have similar color. Traditional machine learning techniques further involve generative and discriminative models to predict the probability of a pixel belonging to skin, which may also take local features like texture into consideration. Even though, these models commonly suffer from low accuracy due to their limited learning capabilities. Early neural network based approaches usually applied Multi-Layer Perceptrons (MLP) whose classification accuracy are still limited. In recent years, fully convolutional neural network (FCN) is widely applied in image segmentation tasks~\cite{long2015fully}, hence skin detection naturally becomes an application of it~\cite{zuo2017combining}. However, the FCN based segmentation usually require large-scale of strong supervision in training stage, which restricts that a high-quality model can be easily trained. In~\cite{tang2018normalized}, a conditional random field is involved as a loss for the end-to-end image segmentation task, which enables the use of weakly supervised data. Unlike these methods, our approach can take advantage of an extra dataset of body segmentation, which is commonly easier to acquire, to boost the performance of a CNN based skin detector.

\paragraph{Multi-task joint learning.}
Multi-task learning (MTL) has been used successfully across all applications of machine learning, from natural language processing~\cite{collobert2008unified} and speech recognition~\cite{deng2013new} to computer vision~\cite{girshick2015fast}. It is generally applied by sharing the hidden layers between all tasks, while keeping several task-specific output layers as branches. Some multi-task networks generally learn common feature maps via shared encoders, so as to potentially improve the performance of all tasks simultaneously. For example, \cite{kendall2018multi} utilized a three-branch network to solve semantic segmentation, instance segmentation and depth prediction in a unified framework. There are more multi-task networks which exist for solving a complicated task, where all the outputs of the task-specified networks are fused for further processing. For example, \cite{wang2018video} proposed a network containing two sub-networks that jointly learning spatial details and temporal coherence for a video inpainting task. In~\cite{han2017high}, Han et al. decompose a shape completion task into two sub-tasks, to reconstruct global and local structure respectively and then fuse together. These methods commonly involve guidance from one branch to another to reduce the learning difficulty. Our approach follows a similar idea, while the two branches of the network can mutually guide the other, so as to boost the performance of skin detection via the recurrent loop in the network.

\section{Algorithm}\label{sec:algo}
Our method is built upon dual-task fully convolutional neural networks. It takes a single RGB image $I$ as input, and produces probability maps of skin $O_S$ and body $O_B$ as outputs. The network contains two decoders $D_S$, $D_B$ in separate branches, for the detection tasks of skin and body respectively. The two decoders share a common encoder $E$, which extracts the feature map of $I$ as $E_I$. The output $O_S$, together with $E_I$, are fed to the decoder of body $D_B$ in the other branch, and vice versa. For either decoder, the output from the other branch serves as a guidance for the task, making the dual tasks mutually guided. The network structure is illustrated at~\figurename~\ref{fig:network-structure}(a).

\subsection{Network with Mutual Guidance}
Our network is a dual-task network with mutual guidance, which can be viewed as a recurrent network due to the structure containing signal loop. For ease of analysis, we decoupled the original network into two stages with no loop as shown in~\figurename~\ref{fig:network-structure}(b). To differentiate the symbols in the two stages, we use $X$ for Stage 1 and $X'$ for Stage 2 accordingly. And for briefness, we use $\kappa\in\{S, B\}$ to represent a module or variable of \textit{skin} or \textit{body}. Here \textit{skin} refers to pixels of entire body skin area, and \textit{body} is a super-set of \textit{skin} that also includes pixels of hair, cloth, etc. A set of $\{X_\kappa\}$ represents both $X_S$ and $X_B$, so that $\{D_\kappa\}$ means $D_S$ and $D_B$ as an example.

In Stage 1, we feed decoders $\{D_\kappa\}$ with guidances $\{G_\kappa\}$ and produce outputs $\{O_\kappa\}$ as intermediate results. Then we feed the decoders with $\{G'_\kappa\}$ in Stage 2 and produce the final outputs $\{O'_\kappa\}$. For two stages, the input $I$ and the weights in $E$ and $\{D_\kappa\}$ are identical, while guidances in two stages are commonly various, i.e. $G_\kappa \neq G'_\kappa$ for $\kappa \in \{S,B\}$. That is because in Stage 1, commonly we have very limited or even no information to provide, while in Stage 2 we have the initial results $\{O_\kappa\}$ detected to serve as guidances. Moreover, here we design a shared encoder $E$ instead of two independent ones, not only for reducing redundancy, but also based on the following two considerations. First, even though the training data for the two tasks have different ground truth, the input RGB images share very similar statistics. Second, there also exist some common properties for the extracted feature map that are desirable for the two tasks, such as robustness to distinguish human foreground and non-human background. Experimental results demonstrate this shared encoder could improve the performance of skin detection by seeing more data and learning the common features, as shown in~\tableautorefname~\ref{tab:values} and~\figurename~\ref{fig:mutual_guidance}. In summary, the entire network can be written as follows.
\begin{align}
\bullet~\ \text{Stage 1}~
\begin{cases}
G_S = e_B, &G_B = e_S \nonumber \\
O_S = D_S(E_I, G_S), &O_B = D_B(E_I, G_B) \nonumber \\
\end{cases} \\
\bullet~\ \text{Stage 2}~
\begin{cases}
G'_S = O_B, &G'_B = O_S \nonumber \\
O'_S = D_S(E_I, G'_S), &O'_B = D_B(E_I, G'_B) \nonumber
\end{cases}
\end{align}
where $e_S$ and $e_B$ are the signals provided as guidances in Stage 1, which are commonly set to 0 in most cases in this paper.
{\color{black}{For the structures of $E$ and $\{D_\kappa\}$, we adapted the stardard UNet~\cite{ronneberger2015u} architecture including 4 downsampling blocks in $E$ and 4 upsamping blocks in $D_\kappa$. The size of input $I$ is $512^2\times3$ so that the feature maps between $E$ and $D_\kappa$, i.e.  $E_I$ is of size $32^2\times1024$. We also applied an encoder of the same structure as $E$ but of half number of channels for each layer to the guidance $\{G_\kappa\}$, to ensure its extracted feature can be well concatenated to $E_I$, after they are fed to $D_\kappa$. For each fully convolutional layer, the kernel size is set to $3\times3$, and is followed by a BatchNorm and a ReLU layer.}}


With the initial results $\{O_\kappa\}$ detected, the decoders are provided with more informative guidances, that are helpful for the second stage detection.

\subsection{Learning Algorithm}\label{problem settings}
The goal of our learning algorithm is to train a dual-task CNN which can detect skin and body end-to-end, which is far from straightforward.
On one hand, for skin detection task, lacking enough training data is a common issue, and human labelling is usually very expensive and time-consuming. On the other hand, for body detection, due to the extensive research in recent years, its data is relatively easier to obtain. So in our problem settings, for each data pair, it contains ground truth mask of skin or body only, noted as $M_S$ or $M_B$. Since there is few such training data triple $(I, M_S, M_B)$ provided, it naturally makes training our network a semi-supervised task, which is achieved by a semi-supervised loss we design and several training details we adopt.

\subsubsection{Semi-supervised loss}
Our newly designed semi-supervised loss consists of three parts, including strongly-supervised and weakly-supervised ones. The former one is the cross-entropy loss between the output and the ground truth; and the latter ones include CRF loss and a weighted cross-entropy (WCE) loss between skin output and body output.

\paragraph{Cross-entropy loss.} As aforementioned, the training data provided to our problem, is a data pair with either skin or body ground truth. For a data sample with $M_S$, we compute the cross entropy losses between $M_S$ and its outputs $O_S$, $O'_S$ respectively, making them strong supervision to the skin detection task. Similarly, it also applies to data sample with $M_B$, so that we produce a sum of four terms of cross-entropy losses:
\begin{align}
\mathbf{L_{ce}} = \sum_{\kappa\in\{S, B\}} \sum_{x\in\{O_\kappa, O'_\kappa\}}l_{\kappa} \cdot L_{ce}(x, M_\kappa)
\end{align}
where $L_{ce}(x,y) = x \cdot \log(y) + (1 - x) \cdot \log(1 - y)$. Here we use a label notation $l_\kappa$ to present whether the current data sample has a ground truth $M_\kappa$. For example, if a data sample has $M_S$ only, then $l_S = 1, l_B = 0$, and vice versa. $l_\kappa$ works as a switch for enabling the contribution of a loss or not. This notation also applies for the rest of this paper.

\paragraph{CRF loss.} For a data sample with a single type of ground truth, one of its outputs can contribute to the cross-entropy loss yet the other one cannot. For this case, we involve a CRF loss as in~\cite{tang2018regularized}. By computing a CRF given image $I$ and a mask $O_\kappa$, CRF loss can constrain neighboring pixels in $I$ with similar color tend to have a consistent label in $O_\kappa$. In most cases when strong supervision is unavailable, this property could potentially refine the output mask. Similarly, the total CRF loss can be written as
\begin{align}
\mathbf{L_{crf}} = \sum_{\kappa\in\{S, B\}} \sum_{x\in\{O_\kappa, O'_\kappa\}}(1 - l_{\kappa}) \cdot L_{crf}(x, I)
\end{align}
where $L_{crf} = S^TWS$ where $W$ is an affinity matrix of $I$ and $S$ is a column vector of flattened $O_\kappa$. We refer the readers to~\cite{tang2018regularized} for more details about CRF loss.

\paragraph{WCE loss.} It is also a prior knowledge that the skin mask should be covered by its body mask for the same image. The consistency should be preserved in the outputs $O_S, O'_S$ and $O_B, O'_B$. For a pixel classified with high probability of skin, it should also have a high probability of body. This does not hold if a lower probability of skin is detected, because the pixel may belong to non-skin regions as cloth or hair, where body probability is still high. To characterize the above relationship, we compute a cross-entropy loss between skin and body probability, then weight it by the skin probability itself, i.e. $L_{wce}(x,y) = x \cdot L_{ce}(x,y)$, where $x\in\{O_S,O'_S\}, y\in\{O_B,O'_B\}$. As a result, the total WCE loss is calculated as
\begin{align}
\mathbf{L_{wce}} = \sum_{\substack{x\in\{O_S,O'_S\}, \ y\in\{O_B,O'_B\}}} L_{wce}(x,y)
\end{align}

\begin{figure}[t!]
  \centering
  \includegraphics[width=0.98\linewidth]{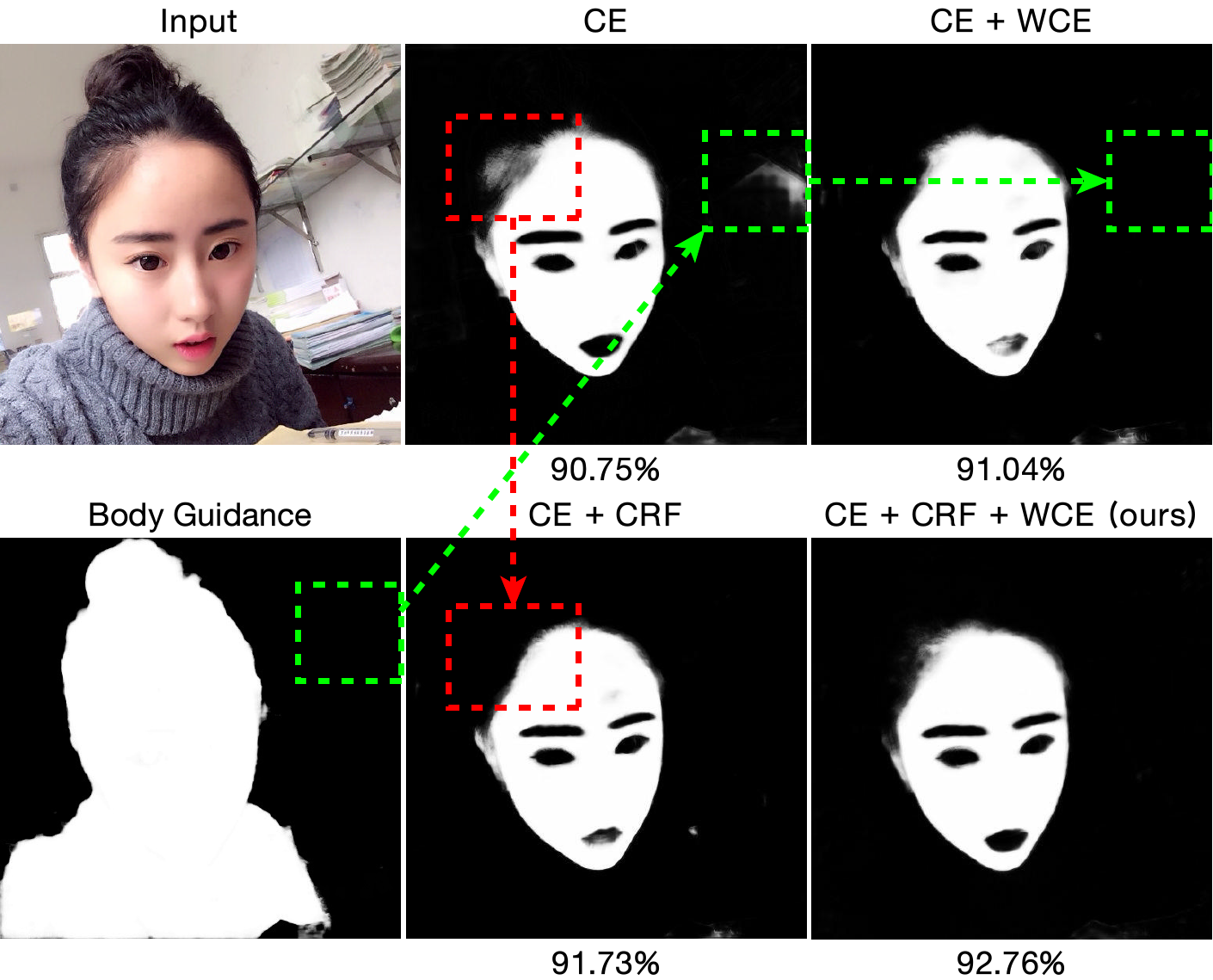}\\
  \caption{Weakly-supervised losses improve the detection result. Red box and arrow: with CRF loss involved, the region between hair and head trends to be classified with the same label, causing the hair region removed. Green boxes and arrows: with body guidance and WCE loss involved, the false alarm region, i.e. the background is removed. The IoUs are listed under the corresponding results. (Best view in color)}\label{fig:image_losses}
\end{figure}

The CRF and WCE are two weakly-supervised losses. Compared with cross-entropy loss as strong supervision, they weakly take effect for the tasks of skin and body detection, which finally improve the performance. To sum up, our semi-supervised loss is
\begin{align}
    \mathbf{L} = \mathbf{L_{ce}} + \lambda_1\cdot\mathbf{L_{crf}} + \lambda_2\cdot\mathbf{L_{wce}}
\end{align}
where $\lambda_1$ and $\lambda_2$ are the balancing hyper-parameters. We set $\lambda_1$ to $0.0001$ and $\lambda_2$ to $0.001$ in our experiments.~\figurename~\ref{fig:image_losses} illustrates an example to reveal the effectiveness of CRF and WCE losses, and more discussion is involved in Section~\ref{subsec:weakly-supervised-losses}.

\begin{table}[t]
\begin{center}
\resizebox{0.99\columnwidth}{!}{
    \begin{tabular}{c
    >{\centering\arraybackslash}p{2.2cm}
    >{\centering\arraybackslash}p{2.2cm}
    >{\centering\arraybackslash}p{2.2cm}
    >{\centering\arraybackslash}p{2.2cm}}
    \toprule
                     & IoU (\%)  & IoU Top-1 (\%) & Precision (\%) & Recall (\%)  \\ \hline
    Thresholding~\cite{kovac2003human}         & 50.84 $/$ {\color{blue}{60.20}} & ~1.06 $/$ {\color{blue}{~0.00}} & 59.30 $/$ {\color{blue}{65.31}} & 81.75 $/$ {\color{blue}{89.58}} \\ \hline
    GMM~\cite{jones2002statistical}                 & 50.06 $/$ {\color{blue}{60.46}} & ~2.34 $/$ {\color{blue}{~0.00}} & 53.45 $/$ {\color{blue}{62.36}} & \textbf{89.31} $/$ {\color{blue}{91.50}}  \\ \hline
    Chen et. al's~\cite{chen2002skin}   & 55.77 $/$ {\color{blue}{62.05}} & ~0.43 $/$ {\color{blue}{~3.12}} & 74.31 $/$ {\color{blue}{72.50}} & 70.94 $/$ {\color{blue}{79.18}} \\ \hline
    Zuo et. al's~\cite{zuo2017combining}   & 69.94 $/$ {\color{blue}{79.81}} & ~0.21 $/$ {\color{blue}{~0.00}} & 84.38 $/$ {\color{blue}{88.97}} & 80.31 $/$ {\color{blue}{88.03}} \\ \hline
    UNet~\cite{ronneberger2015u}                & 75.59 $/$ {\color{blue}{85.50}} & 15.53 $/$ {\color{blue}{28.13}} & 89.38 $/$ {\color{blue}{93.42}} & 83.14 $/$ {\color{blue}{90.91}} \\ \hline
    ResNet50~\cite{he2016deep}            & 75.44 $/$ {\color{blue}{84.33}} & 11.49 $/$ {\color{blue}{12.50}} & 88.77 $/$ {\color{blue}{92.19}} & 82.97 $/$ {\color{blue}{90.72}} \\ \hline
    Deeplab-v3-ResNet50~\cite{chen2018encoder} & 75.97 $/$ {\color{blue}{85.88}} & 10.64 $/$ {\color{blue}{~6.25}} & 86.98 $/$ {\color{blue}{92.51}} & 85.58 $/$ {\color{blue}{\textbf{92.48}}} \\ \hline
    Deeplab-v3-MobileNet~\cite{chen2018encoder} & 73.66 $/$ {\color{blue}{83.96}} & ~7.02 $/$ {\color{blue}{~9.38}} & 87.16 $/$ {\color{blue}{91.91}} & 82.48 $/$ {\color{blue}{90.48}} \\ \hline
    Ours                 & \textbf{81.18} $/$ {\color{blue}{\textbf{87.90}}} & \textbf{51.27} $/$ {\color{blue}{\textbf{40.63}}} &  \textbf{90.01} $/$ {\color{blue}{\textbf{95.23}}} & 89.01 $/$ {\color{blue}{92.08}} \\ 
    \bottomrule%
    \end{tabular}
    } \vspace{5mm}
\resizebox{0.99\columnwidth}{!}{
    \begin{tabular}{c
    >{\centering\arraybackslash}p{2.2cm}
    >{\centering\arraybackslash}p{2.2cm}
    >{\centering\arraybackslash}p{2.2cm}
    >{\centering\arraybackslash}p{2.2cm}}
			\hline
			\toprule
			& IoU (\%)  & IoU Top-1 (\%) & Precision (\%) & Recall (\%)  \\ \hline
			Thresholding~\cite{kovac2003human}        & 50.84 $/$ {\color{blue}{60.20}} & ~3.19 $/$ {\color{blue}{~0.00}} & 59.30 $/$ {\color{blue}{65.31}} & 81.75 $/$ {\color{blue}{89.58}} \\ \hline
			GMM~\cite{jones2002statistical}                 & 50.06 $/$ {\color{blue}{60.46}} & ~5.74 $/$ {\color{blue}{~6.25}} & 53.45 $/$ {\color{blue}{62.36}} & \textbf{89.31} $/$ {\color{blue}{\textbf{91.50}}}  \\ \hline
			Chen et. al's~\cite{chen2002skin}   & 51.44 $/$ {\color{blue}{62.43}} & ~1.28 $/$ {\color{blue}{~6.25}} & 76.11 $/$ {\color{blue}{76.36}} & 63.18 $/$ {\color{blue}{77.89}} \\ \hline
			Zuo et. al's~\cite{zuo2017combining}   & 63.98 $/$ {\color{blue}{73.91}} & ~0.85 $/$ {\color{blue}{~0.00}} & 81.28 $/$ {\color{blue}{85.19}} & 74.99 $/$ {\color{blue}{82.88}} \\ \hline
			UNet~\cite{ronneberger2015u}                & 69.62 $/$ {\color{blue}{79.62}} & 16.81 $/$ {\color{blue}{18.75}} & 83.96 $/$ {\color{blue}{89.55}} & 80.61 $/$ {\color{blue}{87.87}} \\ \hline
			ResNet50~\cite{he2016deep}            & 66.03 $/$ {\color{blue}{77.97}} & 7.66 $/$ {\color{blue}{~3.12}} & 84.73 $/$ {\color{blue}{88.30}} & 74.82 $/$ {\color{blue}{86.87}} \\ \hline
			Deeplab-v3-ResNet50~\cite{chen2018encoder}  & 69.04 $/$ {\color{blue}{76.63}} & 12.34 $/$ {\color{blue}{12.50}} & 81.81 $/$ {\color{blue}{86.19}} & 81.34 $/$ {\color{blue}{87.39}} \\ \hline
			Deeplab-v3-MobileNet~\cite{chen2018encoder} & 67.95 $/$ {\color{blue}{77.63}} & ~6.60 $/$ {\color{blue}{~9.38}} & 81.92 $/$ {\color{blue}{86.93}} & 79.90 $/$ {\color{blue}{87.59}} \\ \hline
			Ours                 & \textbf{75.29} $/$ {\color{blue}{\textbf{81.89}}} & \textbf{45.53} $/$ {\color{blue}{\textbf{43.75}}} &  \textbf{87.34} $/$ {\color{blue}{\textbf{92.58}}} & 84.64 $/$ {\color{blue}{87.51}} \\ 
			\bottomrule%
		\end{tabular}
}
\end{center}
    \vspace{-1.5em}
    \caption{Evaluated IoU, IoU Top-1 rate, precision and recall on our validation dataset (black) and Pratheepan Face dataset ({\color{blue}{blue}}), trained by balanced dataset ($\# \textit{skin}, \# \textit{body} = 5k$) (top) and unbalanced dataset ($\# \textit{skin} = 1k, \# \textit{body} = 5k$) (bottom).}\vspace{-0.6em}\label{tab:values}
\end{table}
\subsubsection{Training details}
\paragraph{Dual-task joint learning.} Our network is trained by the Adam Optimizer, where each branch is handled exclusively in each iteration, while they are jointly learned for the dual-task. For even and odd iteration, we feed data samples with $M_S$ and $M_B$ respectively, i.e. $(I,M_S,M_B=0,l_S=1,l_B=0)$ or $(I,M_S=0,M_B,l_S=0,l_B=1)$. Given each data sample, thanks to the existence of label $l_\kappa\ (\kappa\in\{S,B\})$, its cross-entropy loss is computed in one branch and CRF loss is done in the other. With the training going on, the outputs from Stage 1 $\{O_\kappa\}$, gradually provide guidance for Stage 2. Meanwhile, with the increasingly informative guidances from $\{O_\kappa\}$, the detection difficulty for decoders in Stage 2 is reduced, so that the final outputs $\{O'_\kappa\}$ are expected to become increasingly accurate.

\paragraph{Finetune.} To develop the potential of the dual-task network with mutual guidance, care must be taken during training. In practice, we first train the Stage 1 network by involving losses on $\{O_\kappa\}$ only. Due to the lack of guidance at present stage, we feed the network with $G_\kappa = E_\kappa = 0, \kappa\in\{S,B\}$ instead. With the convergence of the network, the outputs $\{O_\kappa\}$ tend to become informative but still of limited accuracy.

We further involve training in the 2nd stage. We feed the decoders $\{D_\kappa\}$ with $\{G'_\kappa\}$, where $G'_\kappa$ is obtained with the following manner. For a data sample with $M_S\neq0$ and $l_S=1$, $G'_B$ is set to $M_S$; otherwise, $G'_B = O_S$. Similar rules also apply to $G'_S$, i.e.
\begin{align}
\begin{cases}
G'_B = l_S \cdot M_S + (1 - l_S) \cdot O_S \nonumber \\
G'_S = l_B \cdot M_B + (1 - l_B) \cdot O_B
\end{cases}
\end{align}

This strategy ensures that, we feed the most trusted data as guidance to the decoders, to avoid misleading with incorrect guidance data, especially if $O_\kappa$ is of low quality. Furthermore, due to the large variation between guidances in the two stages, i.e. $G_\kappa \neq G'_\kappa$, the decoders of sharing weights in the two stages have to own the power to regress the same data sample with various guidance, i.e. $(I,G_S,G_B)$ and $(I,G'_S,G'_B)$, to the unique ground truth $M_S$ (if $l_S = 1$) or $M_B$ (if $l_B = 1$). To achieve it, we apply a gradient stopping scheme to disable the back-propagation from $G'_B, G'_S$ to their corresponding decoders in Stage 1, so as to avoid the outputs $\{O_\kappa\}$ tending to trivially regress to values like $\{E_\kappa\}$. Meanwhile, the semi-supervised loss additionally involves the ones computed with $\{O'_\kappa\}$. With the training keeping on, the decoders gradually obtain the tolerance to handle various guidances in the two stages, while with informative guidance they can perform better. We demonstrate the effectiveness of the two-stage training strategy, mutual guidance and gradient stopping scheme in Section~\ref{sec:ablation}.

\begin{figure}[t!]
    \centering
    \includegraphics[trim=33mm 6mm 40mm 8mm, clip, width=0.98\linewidth]{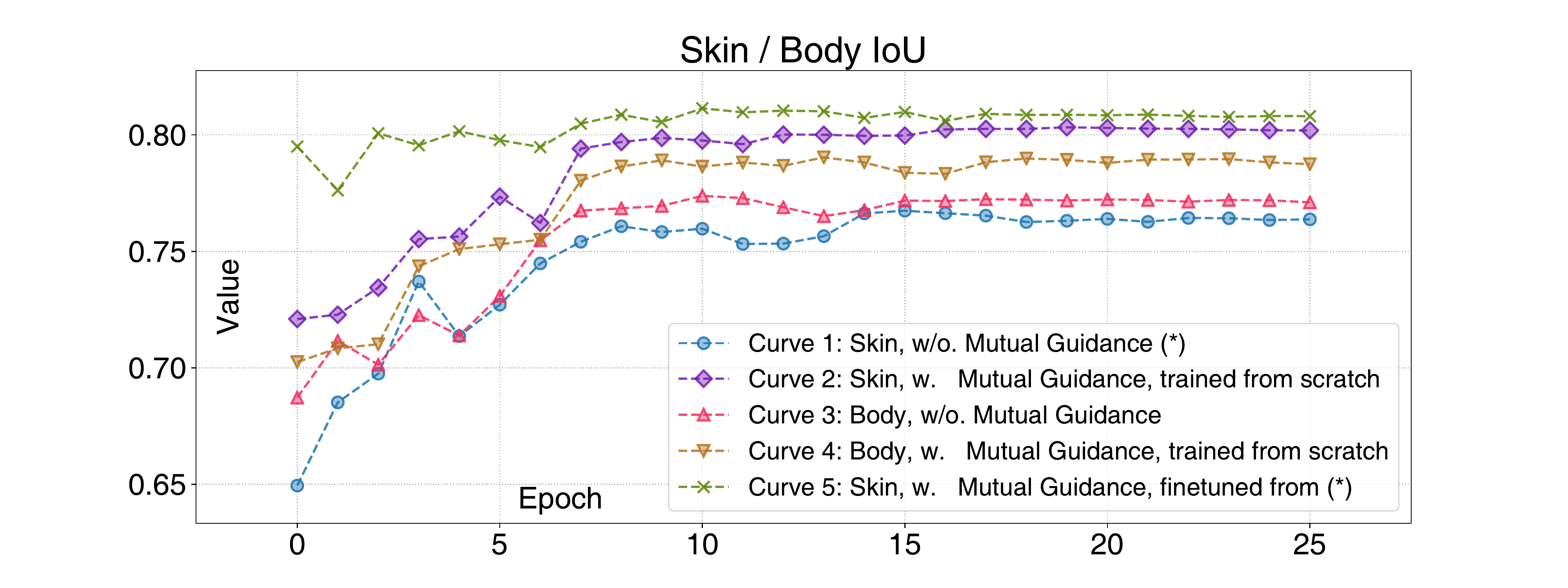}\\
    \includegraphics[width=0.98\linewidth]{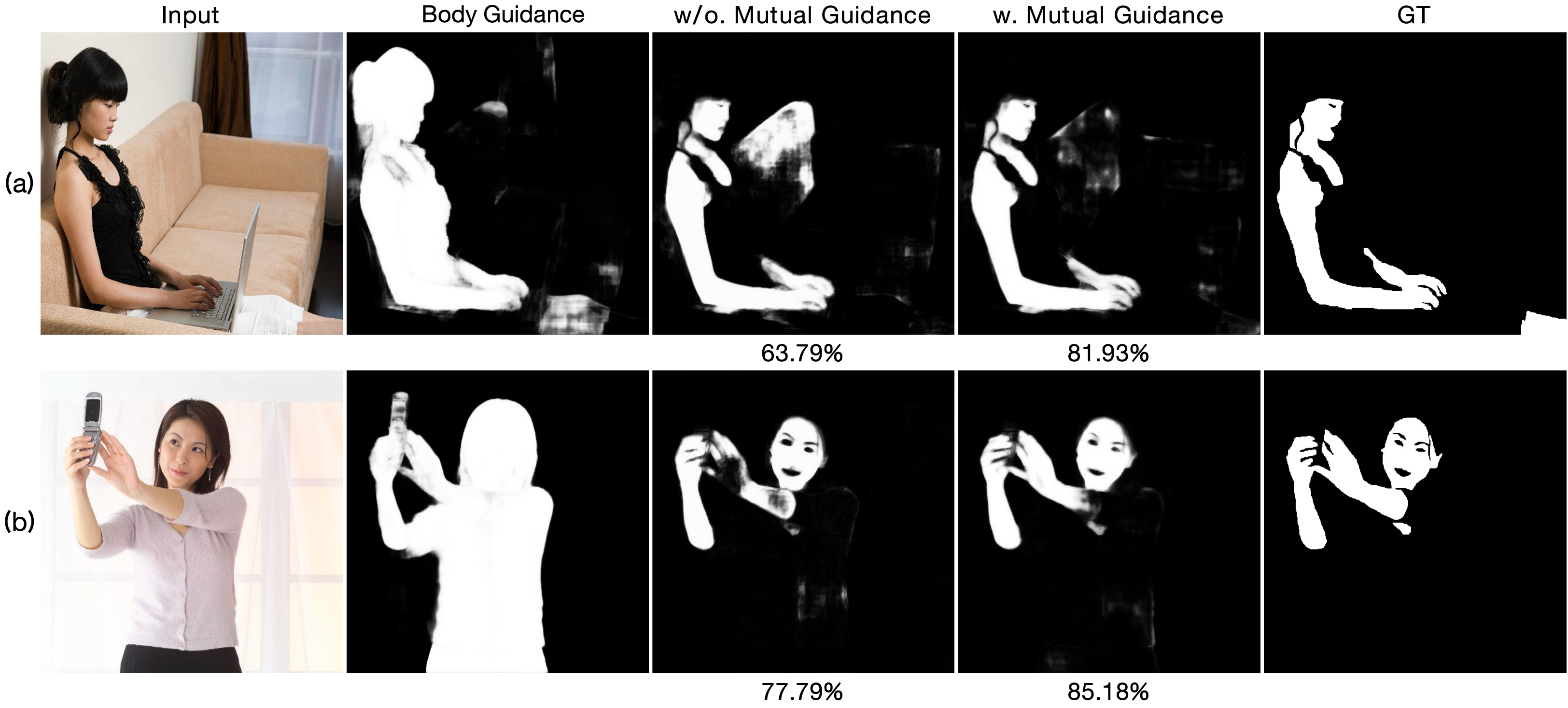}
    \caption{Mutual guidance. Top: Curve $1\sim4$: IoU for detected skin and body mask by our dual-task network, trained with or without mutual guidance, with respect to the number of epoches. Curve $5$: detected skin IoU by our network in finetuned version. Bottom: two examples showing the detected masks by our network without (column 3) or with (column 4) mutual guidance. The body masks serving as guidances are shown in column 2. }
    \label{fig:mutual_guidance}
\end{figure}

\begin{figure*}[t]
    \centering
    \includegraphics[width=1.0\linewidth]{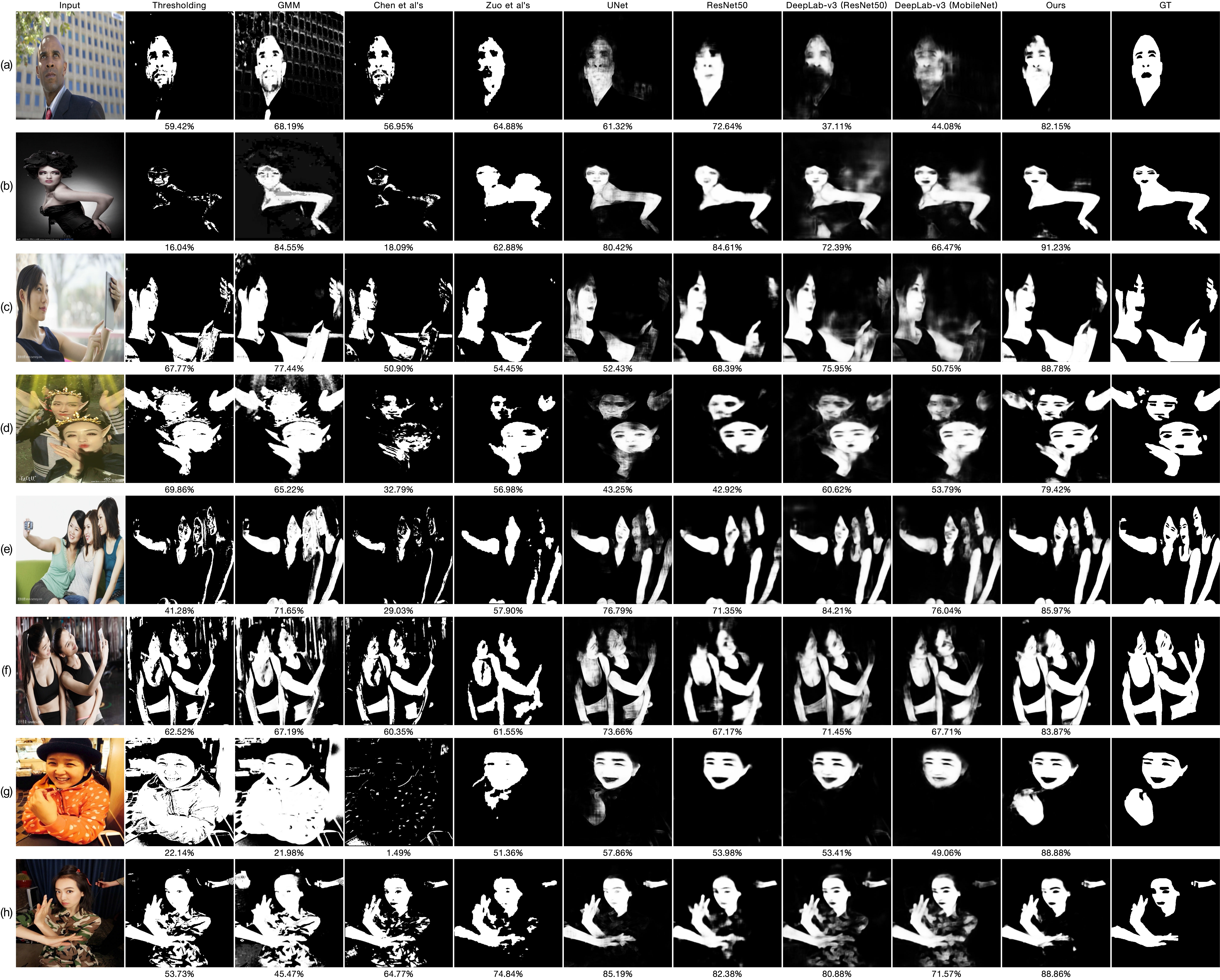}
    \caption{Typical skin detection results on our validation dataset, by various methods including thresholding, GMM, Chen et al's~\cite{chen2002skin}, Zuo et al's~\cite{zuo2017combining}, UNet, ResNet50, DeepLab-v3 with ResNet50 as backbone, DeepLab-v3 with MobileNet as backbone, ours (Column 2 to 10). Input and ground truth are shown in column 1 and 11. }
    \label{fig:image-results}
\end{figure*} 

\section{Experimental Results}\label{sec:results}
\subsection{Dataset and Implementation Details}
{\color{black}{Our dataset is composed of 10,711 RGB images, 5,000 of which have human-annotated skin masks $M_S\ (l_B = 0, l_S = 1)$ and the rest have body masks $M_B\ (l_S = 0, l_B = 1)$, noted as $\mathcal{D_S}$ and $\mathcal{D_B}$. The original RGB images are collected from the Internet, and we resized them into $512^2$ resolution. We randomly selected 470 samples from $\mathcal{D_S}$ and 475 ones from $\mathcal{D_B}$, to establish two validation datasets. During training, we augmented the training data by randomly flipping, resizing and cropping the original data samples to ensure data diversity. Our code was developed with TensorFlow, and the whole training was completed in about 12 hours by one NVIDIA GeForce GTX 1080Ti GPU. We will release our dataset to the public upon the acceptance of this paper.
\begin{figure*}[t!]
  \centering
    \includegraphics[trim=9mm 12mm 19mm 16mm, clip, width=0.33\linewidth]{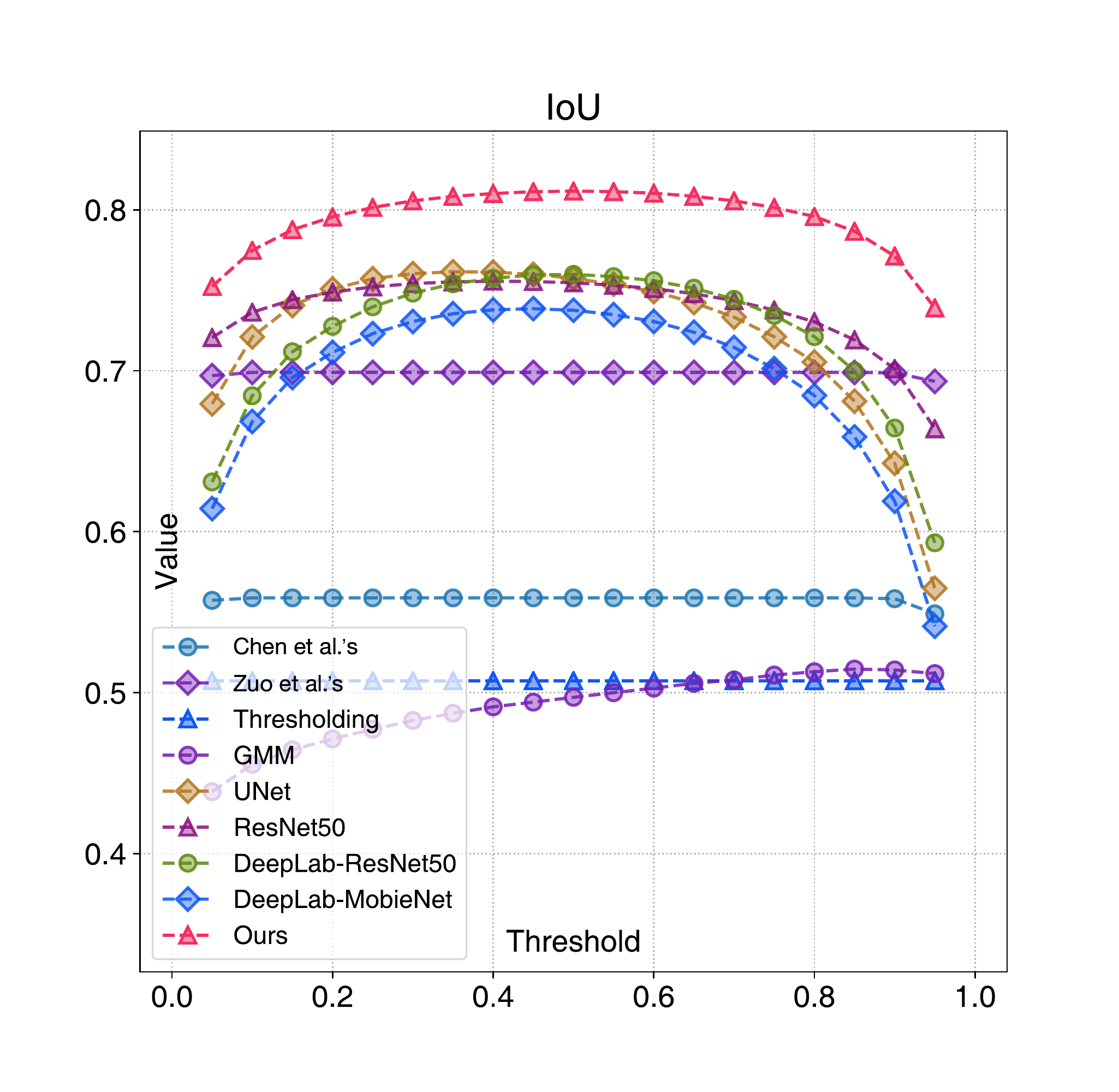}
    \includegraphics[trim=9mm 12mm 19mm 16mm, clip, width=0.33\linewidth]{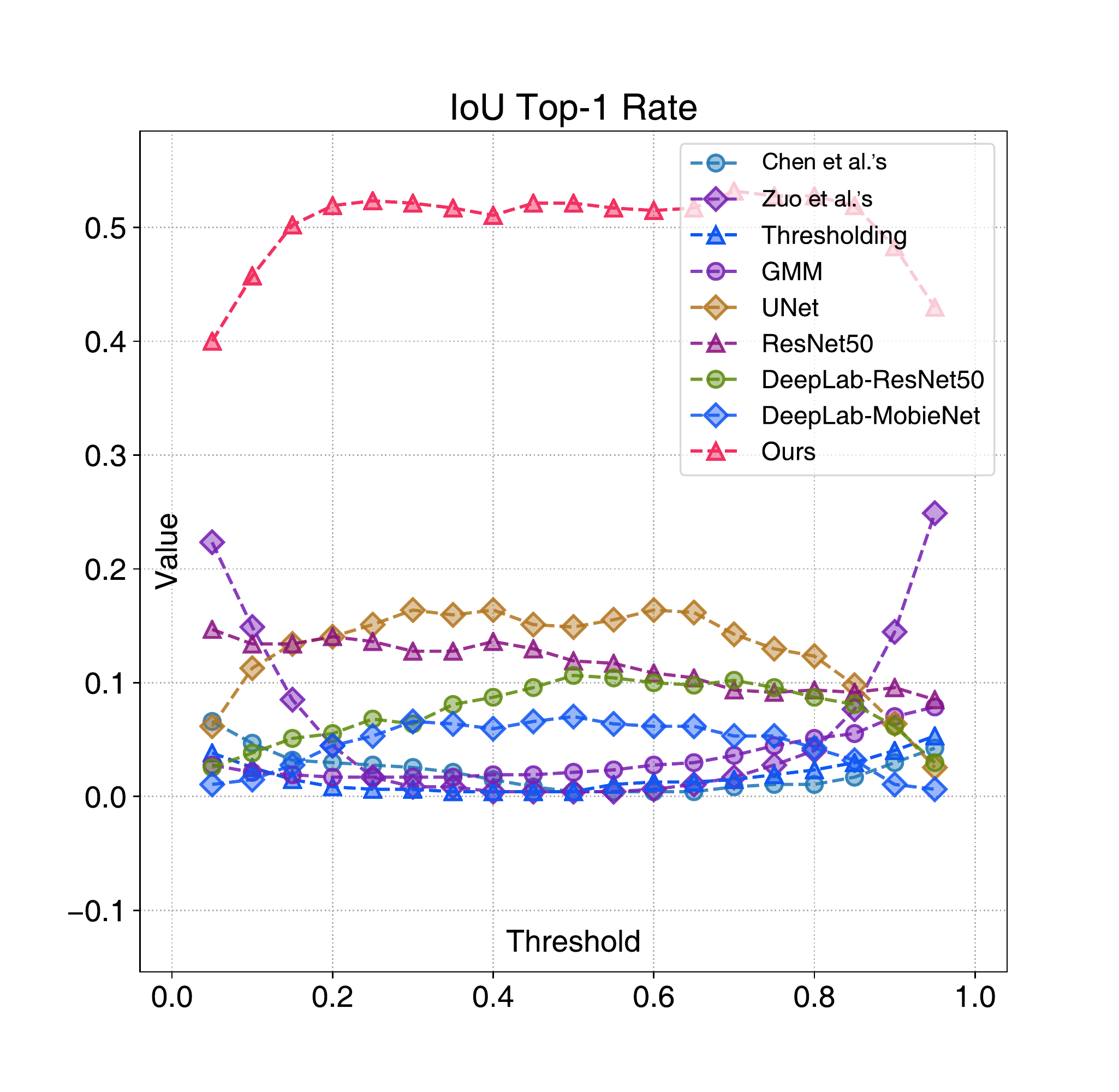}
    \includegraphics[trim=9mm 12mm 19mm 16mm, clip, width=0.33\linewidth]{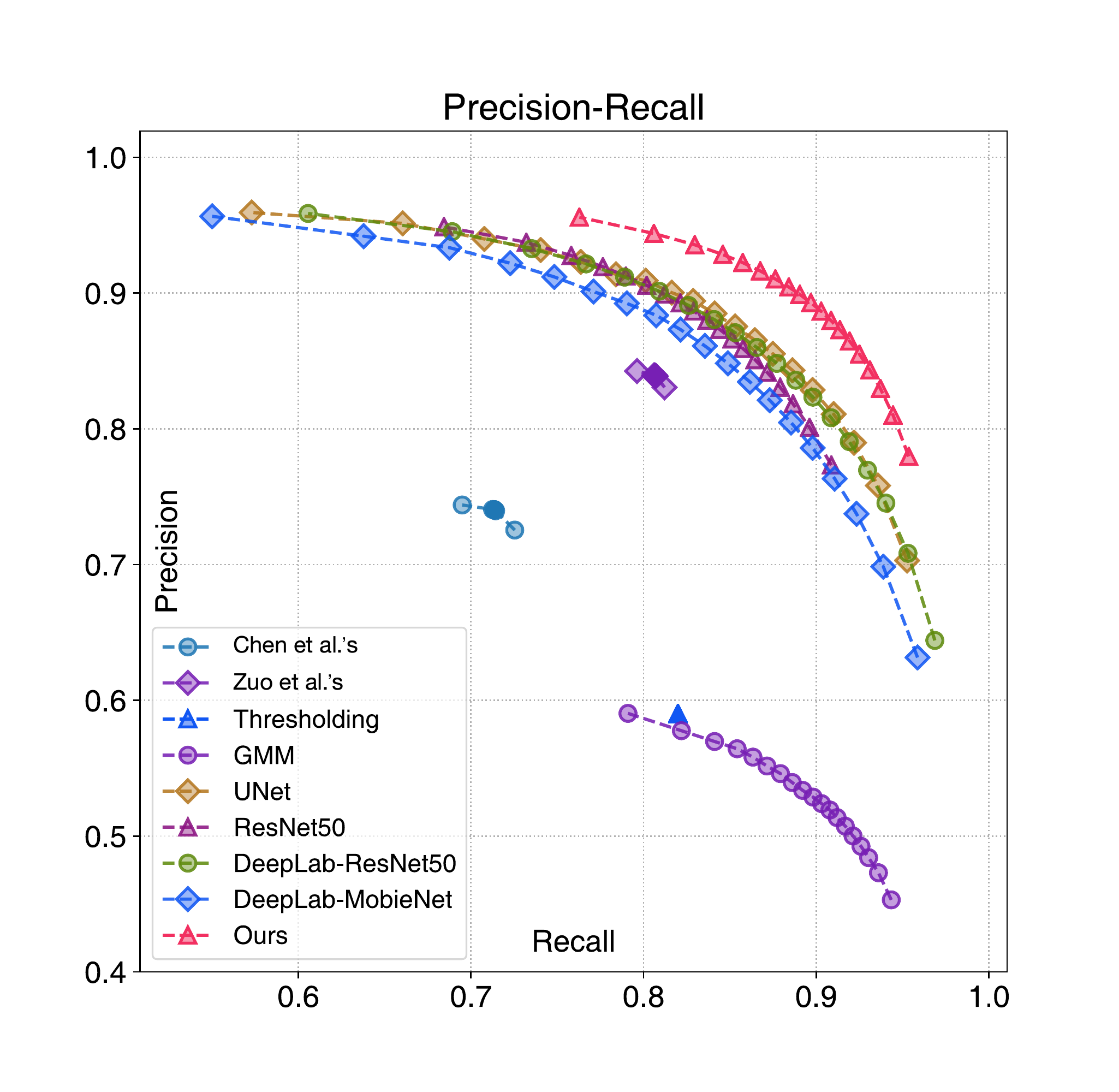}

  \caption{Curves of IoU, IoU Top-1 rate with respect to the probability threshold, and Precision-Recall on our dataset.}\label{fig:roc-curves}
\end{figure*}

}}

\subsection{Comparison with Existing Methods}
We compared our method with some state-of-the-art ones, including two traditional algorithms and six NN based methods. ~\cite{kovac2003human} is a pixel value thresholding method which establishes some rules on pixel RGB and HSV color to classify a pixel to skin or not, rather than a soft probability map. Gaussian Mixture Model (GMM)~\cite{jones2002statistical} based method improves the mechanism, where a skin color GMM is learned, given an initial skin mask. The learned GMM then predicts the skin probability for each pixel. The problems behind the two traditional methods are there lacks high level features involved in the detection task, and they are far from robustness to light change or complex background. The other six NN based methods are end-to-end, producing a skin probability map given an RGB image, where the differences lie in the structures of networks only, i.e. Chen et al's~\cite{chen2002skin}
, Zuo et al's~\cite{zuo2017combining}, U-Net~\cite{ronneberger2015u}, ResNet50~\cite{he2016deep} and DeepLab-v3~\cite{chen2018encoder} with ResNet50 or MobileNet as backbones.

We trained the six networks to convergence with multiple trials with dataset $\mathcal{D_S}$, and selected their best results. To quantitatively compare our method with them, we evaluated precision, recall, and intersection-over-union (IoU) of all the results, and list them at~\tableautorefname~\ref{tab:values}. The data shows that in terms of IoU and precision, our approach outperforms the state-of-the-art for skin detection. For recall, our method ranks only below the GMM method, which has more false alarms so as to suffer from a poor precision. For the mean IoU of all the validation data samples, our method is about 4\% higher than the 2nd competitor in average. Even though, we further evaluates the robustness of our method by calculating an IoU Top-1 rate, i.e. what percentage of data each method can win the competition in terms of IoU. We found our method wins for nearly 51\% validation data and none of the others has a comparable performance. We illustrate the curves of IoU, IoU Top-1 rate and precision-recall at~\figurename~\ref{fig:roc-curves}.  
We also compared our network with four CNNs on the public dataset of Pratheepan Face~\cite{tan2012gesture}, and results also show that our method outperforms the others in~\tableautorefname~\ref{tab:values} (blue values).

We list several typical detected skin masks in~\figurename~\ref{fig:image-results} and~\ref{fig:teaser} for qualitative comparison, where the examples cover various skin colors, complex illuminance, white balance, similar color in background especially cloth etc. They are captured in various conditions, by casual cameras or in studio. ~\figurename~\ref{fig:image-results}(a) is a man of black skin wearing a navy suit, and (h) captures an asian girl wearing a camouflage suit with spots in skin color, making their skins so hard to distinguish; (b) contains white background light around the woman's naked back and arm, and (g) is in a warm color style; (d)(e)(f) contain multiple people in various poses, especially in (d) three people (2 close, 1 far away, with various scales) exist in a yellow lighting condition. (c) shows a woman holding a phone which has reflectance and occludes part of her arm, making the visible skin spatially discontinuous. These challenging conditions make other methods fail or perform poorly, for example traditional thresholding or GMM method totally fail in (g)(h) and the end-to-end CNN methods work unstable in (a)$\sim$(f). In contrast, our approach overcomed the difficulties as stated above and produced accurate and robust results, especially in~\figurename~\ref{fig:image-results}(d), where the man in the distance looks too tiny to be visible for human eyes.



\subsection{Ablation Studies}\label{sec:ablation}
\subsubsection{Mutual guidance}
We further reveal the effectiveness of mutual guidance scheme by experiments with or without it, both trained from scratch for fair comparison. By disabling the mutual guidance, i.e. training the proposed dual-task network in Stage 1 only, we plot the IoU of skin and body in the validation dataset for every epoch until convergence, as illustrated in~\figurename~\ref{fig:mutual_guidance}. From it, we can see that with mutual guidance involved, the IoU for both skin and body can be raised to a higher value at the same epoch, compared with the case of mutual guidance excluded. Note that, even for the case without mutual guidance, our network achieves $\mathbf{76.74}\%$ IoU, still higher than state-of-the-art single-task CNN solutions as shown in~\tableautorefname~\ref{tab:values} (top, black values). It is due to the structure of our network with shared encoder $E$, which enables the learning from the extra body data. We also show two skin detection results by the two methods in~\figurename~\ref{fig:mutual_guidance}(a)(b) for a visual comparison. In both examples, the network without mutual guidance produced results with lower IoU due to false alarm (sofa in (a)) or mis-detection (hand in (b)). With body guidance involved, performance is improved with the false positive pixels and mis-detected pixels being corrected.

\subsubsection{Weakly supervised losses}\label{subsec:weakly-supervised-losses}
We also demonstrate the effectiveness of the weakly supervised losses we introduce, by disabling either one of them or both of them. We found that although these two losses contribute insignificantly compared with the strong supervised cross-entropy loss, they indeed take effects proven by the fact that each one raises up IoU by approximate 0.25\%, and both can raise up to 1.9\%, as shown in~\tableautorefname~\ref{tab:weakly-losses} Top.~\figurename~\ref{fig:image_losses} illustrates an example where if neither of the CRF and WCE losses is involved, there exists some misclassified background pixels. In this case, WCE takes effects because the detected body mask supervises the region to be classified as background. Meanwhile, CRF loss weakly supervises the region between the hair and head to have a consistent labeling, causing the hair pixels filtered out. With both losses enabled, the final IoU outperforms the CE-loss-only version by 2\% .

\subsubsection{Unbalanced dataset}
We also conducted a comparison on unbalanced dataset. In this experiment, we extracted only $1k$ skin samples from $\mathcal{D_S}$, together with the $5k$ body samples $\mathcal{D_B}$ for training. We also list the IoU, IoU Top-1, Precision and Recall in Table~\ref{tab:values} (bottom). Compared with the results trained by balanced dataset, the IoU value drops about 6\% for our method but is still obviously higher than the others. We also applied this experiment to Pratheepan Face dataset~\cite{tan2012fusion} and similar conclusion was drawn.

\begin{table}[t]
    \centering
    \resizebox{0.97\columnwidth}{!}{
    \begin{tabular}{
    >{\centering\arraybackslash}p{2.6cm}
    >{\centering\arraybackslash}p{2.8cm}
    >{\centering\arraybackslash}p{2.4cm}
    >{\centering\arraybackslash}p{2.2cm}}
    \toprule
            CE (Strong) & CRF & WCE & IoU (\%) \\ \hline
$\checkmark$ &  &                           & 79.28 \\ 
$\checkmark$ & $\checkmark$ &               & 79.48 \\ 
$\checkmark$ &  & $\checkmark$              & 79.52 \\ 
$\checkmark$ & $\checkmark$ & $\checkmark$  & 81.18  \\ \bottomrule
    \end{tabular}}\\ \vspace{1pt}

    \resizebox{0.97\columnwidth}{!}{
    \begin{tabular}{
    >{\centering\arraybackslash}m{2.6cm}
    >{\centering\arraybackslash}m{2.8cm}
    >{\centering\arraybackslash}m{2.4cm}
    >{\centering\arraybackslash}m{2.2cm}}
    \toprule
          & Ours (DeepLab-v3 -MobileNet) (\%)& Ours (UNet) (\%) & IoU Gain by Method (\%) \\
    \hline
    w/o. M.G.         & 75.15  & 76.56  &  $\uparrow$~1.41  \\
    w. M.G.           & 79.02  & 80.11  &  $\uparrow$~1.09  \\
    IoU Gain by M.G.  & $\uparrow$~3.87   & $\uparrow$~3.55   &  $-$\\
    \bottomrule
    \end{tabular}%
    }\vspace{2pt}
    \caption{Performance for various compositions of losses (top) and different backbone networks (bottom). M.G. is the abbreviation of Mutual Guidance.}
    \label{tab:weakly-losses}\vspace{-3mm}
\end{table}



\subsubsection{Backbone networks}
We also explore the influence by the backbone network embedded in our network structure, by replacing the existing U-Net structure with a DeepLab-v3 with MobileNet backbone structure, whose number of parameters is about 60\% of UNet. Experimental results show that, in this smaller network, lower IoU is obtained but more capability of mutual guidance is released. See~\tableautorefname~\ref{tab:weakly-losses} bottom for more comparison details. \vspace{-2mm}


\subsubsection{Training strategy}
\begin{figure}[t!]
  \centering
  \includegraphics[width=0.98\linewidth]{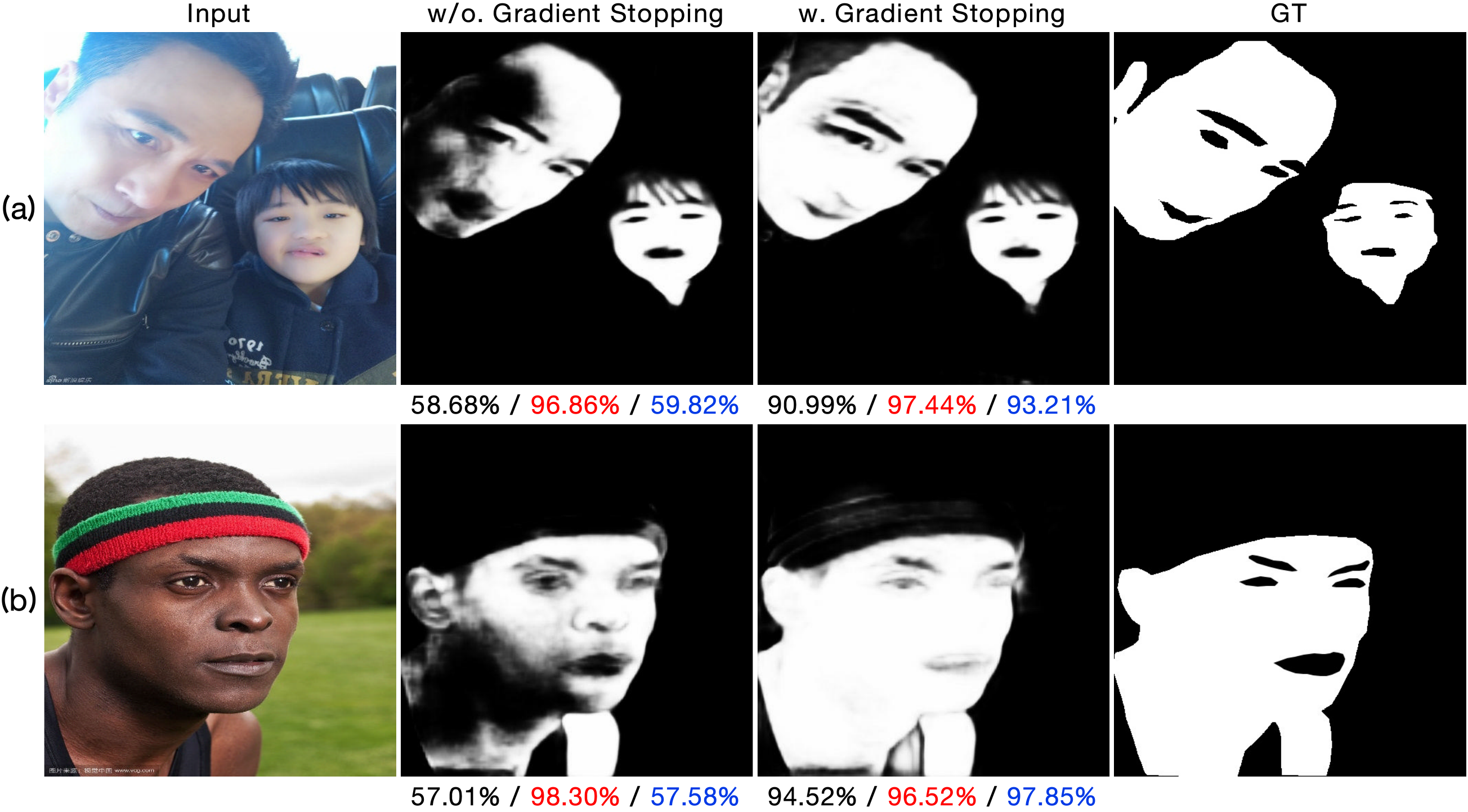}\\
  \caption{Skin detection results trained without (Column 2) or with (Column 3) gradient stopping. The three values under the results are IoU (black), Precision (red) and Recall (blue).}\label{fig:gradient_stop}\vspace{-3mm}
\end{figure}

\paragraph{Gradient stopping.}
We also conducted an experiment to check the necesity of gradient stopping.~\figurename~\ref{fig:gradient_stop} shows two examples. From them, we see that with gradient stopping disabled, the detected skin masks tend to have a high precision but low recall, which is more likely to be trivial results like $e_\kappa = 0$. This is a local minimum of our network, caused by the setting $G_\kappa = e_\kappa = 0$ in Stage 1. When gradient stopping is enabled, we keep the gradients from being back-propagated to $\{O_\kappa\}$, so that the trivial local minimum cannot be easily reached. \vspace{-2mm}
\paragraph{Initial guidance $e_S, e_B$.}
We also conducted an experiment by providing the guidance $\{G_\kappa\}$ with $\{e_\kappa\neq 0\}$. Specifically, $e_B$ is a body bounding box mask and $e_S$ is skin detection results by GMM. We also trained our network with mutual guidance from scratch, and achieved 80.74\% IoU. This value is higher than 80.11\% which was produced by the version of $\{e_\kappa = 0\}$, meaning by providing more informative guidance in Stage 1, our network could be more easily trained. \vspace{-2mm}

\paragraph{Finetune.} We also compared the performance of our network between train-from-scratch and finetune versions, illustrated in Curve 2 and 5 in~\figurename~\ref{fig:mutual_guidance} Top. With finetune involved, our network obtained a higher average IoU in validation dataset.

\section{Conclusion}\label{sec:conclu}
We have presented a new data-driven method for robust skin detection from a single human portrait image. To achieve this goal, we designed a dual-task neural network for joint detection of skin and body. Our dual-task network contains a shared encoder but two decoders, for the two tasks separately. The two decoders work in a mutually guided manner, i.e. either output of the skin or body decoder also serves as a guidance to boost the detection performance for its counterpart. Furthermore, our network can be trained in a semi-supervised manner, i.e. we do not require both types of ground truth exist in one training data sample. It is achieved by a newly designed semi-supervised loss as proposed. We conducted extensive experiments to demonstrate the effectiveness of mutual guidance, semi-supervised losses and various training strategies. Results also show that our method outperforms the state-of-the-art in skin detection. We also hope that the idea of mutual guidance could inspire more works in related problems like image/video denoising~\cite{liu2019learning}, detection~\cite{ren2016look},  completion~\cite{ding2019frame,yamaguchi2018high,huynh2018mesoscopic}, segmentation~\cite{wang2014video}, generation or compression~\cite{wang2019gif2video,qiu2019two,meng2018mganet,wang2017video,zhou2018hairnet} etc. in the future.

\section*{Acknowledgements}
This work is partially supported by the National Natural Science Foundation of China under Grant No. 61672125, No. 61702565 and No. U1811463. We also thank the anonymous reviewers' comments to facilitate the improvement of this paper.

{\small
\bibliographystyle{ieee}
\bibliography{egbib}
}

\end{document}